\RequirePackage{fix-cm}
\documentclass[twocolumn]{svjour3}          % twocolumn
\smartqed  % flush right qed marks, e.g. at end of proof

\usepackage{framed}
\usepackage{amssymb}
\usepackage{marvosym}
\usepackage{graphicx}
\usepackage[numbers]{natbib}
\usepackage{enumitem}
\usepackage{chngpage} %it shouldn't be needed for a final version
\usepackage{bm} %it shouldn't be needed for a final version
\newenvironment{algo}{\framed\noindent\textbf{\textsf{Algorithm:}}}{\endframed}

\begin{document}

\title{Hybrid Focal Stereo Networks for Pattern Analysis in Homogeneous Scenes%\thanks{Grants or other notes
%about the article that should go on the front page should be
%placed here. General acknowledgments should be placed at the end of the article.}
}
%\subtitle{Do you have a subtitle?\\ If so, write it here}

%\titlerunning{Short form of title}        % if too long for running head

\author{Emanuel Aldea         \and
        Khurom H. Kiyani %etc.
}

%\authorrunning{Short form of author list} % if too long for running head

\institute{E. Aldea (\Letter) $\cdot$ K. H. Kiyani \at
              Communications and Signal Processing Group, Dept. of Electrical and Electronic Engineering, Imperial College London, South Kensington, London SW7 2AZ, United Kingdom \at Systems and Applied Physics Group, AquaMed Research and Education, Doha, Qatar \\
              Tel.: +33-648936199\\
              Fax:  +44-2075945017\\
              \email{e.aldea@imperial.ac.uk}           %  \\
%             \emph{Present address:} of F. Author  %  if needed
           \and
           K. H. Kiyani \at
              \email{k.kiyani@imperial.ac.uk}
}

\date{Received: date / Accepted: date}
% The correct dates will be entered by the editor

\maketitle

\begin{abstract}
In this paper we address the problem of multiple camera calibration in the presence of a homogeneous scene, and without the possibility of employing calibration object based methods. The proposed solution exploits salient features present in a larger field of view, but instead of employing active vision we replace the cameras with stereo rigs featuring a long focal analysis camera, as well as a short focal registration camera. Thus, we are able to propose an accurate solution which does not require intrinsic variation models as in the case of zooming cameras. Moreover, the availability of the two views simultaneously in each rig allows for pose re-estimation between rigs as often as necessary. The algorithm has been successfully validated in an indoor setting, as well as on a difficult scene featuring a highly dense pilgrim crowd in Makkah. 
%\keywords{First keyword \and Second keyword \and More}
% \PACS{PACS code1 \and PACS code2 \and more}
% \subclass{MSC code1 \and MSC code2 \and more}
\end{abstract}

\section{Introduction}
\label{sec:intro}
The problem of multiple camera calibration, or camera network calibration, has been a central topic for the pattern recognition and robotics communities since their inception. Moreover, the use of camera networks has become pervasive in our society; beside their use in surveillance and security enforcement, cameras are heavily relied upon in application domains related to entertainment and sports, geriatrics and elderly care, the study of natural and social phenomena, etc. Motivated by all these developments, a large body of work has been devoted to the problem of estimating accurately the camera network topology, i.e. camera positions and orientations in a common reference system. Inferring the topology in camera networks with non-overlapping fields of view (FOV) is a topic specific to wide-area tracking relying more on high-level image processing and statistical inference and will not be addressed in the current work; the focus of the current article is on estimating the geometric topology for cameras with overlapping FOV. Although such a network may be composed of a large number of cameras firmly attached to a mobile object such as a robot, car, or UAV, most commonly camera networks are static and point towards a specific scene of interest. In these cases, multiple camera calibration is performed by using a specific calibration pattern or object \citep{Zhang98aflexible, Baker00, Svoboda2005}, which is deployed and moved in the scene during a dedicated calibration phase. If the use of a calibration object is not possible, scene based calibration may be performed by exploiting visible interest points in methods based on pose refinement \citep{Triggs99}, or if applicable by using dynamic silhouettes, such as in \cite{poll10}.

The analysis of a homogeneous scene which is not accessible, or where using calibration objects is not feasible, raises a problem which is not solved by the common methods used for multiple camera calibration. We approach this problem by replacing cameras with hybrid static stereo rigs, where a long focal camera is used for analysis and a large FOV camera is used for registration with other rigs. The large FOV cameras do use salient features of the larger scene in order to perform relative pose estimation, but in relying on this relatively straightforward solution we avoid using active cameras which require complex models for the dynamic evolution of their intrinsic parameters. Other benefits of possessing simultaneous large and small FOV images of the scene are the fact that the registration does not assume anything about the analysed scene, the fact that the salient features do not have to be static as long as the cameras are accurately synchronized, but then if they are static they can be used to re-estimate continuously the pose and correct phenomena such as camera shaking. Although our aim and the experiments in the current work are related to pattern analysis in highly dense crowds, we hope that the proposed algorithm may be useful in a variety of applications requiring accurate analysis of homogeneous scenes inaccessible for calibration.  

The outline of the paper is as follows. In Section \ref{sec:related} we illustrate the fundamental problem that we address, and discuss related work and alternative solutions. Then, Section \ref{sec:background} recalls the fundamental notions which are required for scene based calibration and for the understanding of the proposed algorithm, which is presented in Section \ref{sec:algo}. Section \ref{sec:results} illustrates an application of the proposed algorithm to the analysis of a highly crowded scene, and Section \ref{sec:conclusion} presents the conclusions.

\label{sec:related}

\subsection{ FOV choice and need for hybrid stereo}
\label{subsec:stereo}
\begin{figure}[thpb]
      \centering
      \includegraphics[width=0.47\textwidth]{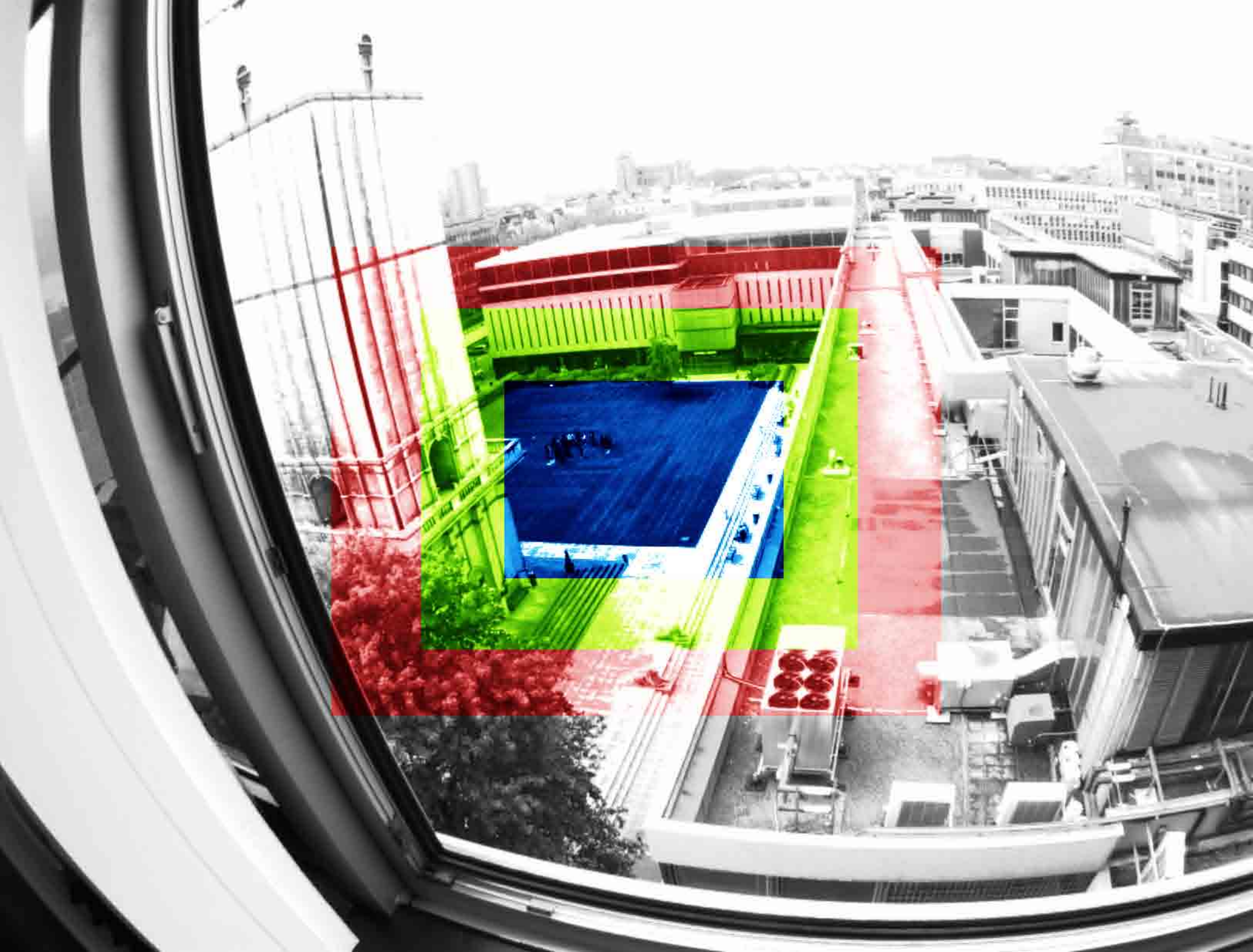}
      \caption{Differents FOV superposed approximately on the image corresponding to focal $f = 4mm$ for a visual comparison. The FOV are highlighted as : red for $f = 8$ mm; green for $f = 12$ mm; blue for $f = 16$ mm. }
      \label{fig:comp}
\end{figure}

Based on the simple pinhole projection model (also recalled in Section \ref{subsec:model}), let us illustrate an issue related to the representation of a homogeneous region of interest in a camera sensor (for all the following tests and examples we will employ Sony ICX274 sensors with a 8.923 mm diagonal and an effective pixel resolution of 1624 $\times$ 1234). In Figure \ref{fig:comp}, we provide a visual comparison between the FOV of a 4mm lens and the FOV of lenses with progressively higher focal lengths (8, 12 and 16 mm), superposed on the initial image. The relative comparison highlights the fact that the scarcity of salient features increases dramatically as the FOV focuses on the central interest area. Consequently, this has an immediate impact on the feasibility and robustness of relative pose estimations between this view and other possible views aimed from different positions at the same area.

\begin{figure*}[htpb]
      \centering
      \includegraphics[width=0.9\textwidth]{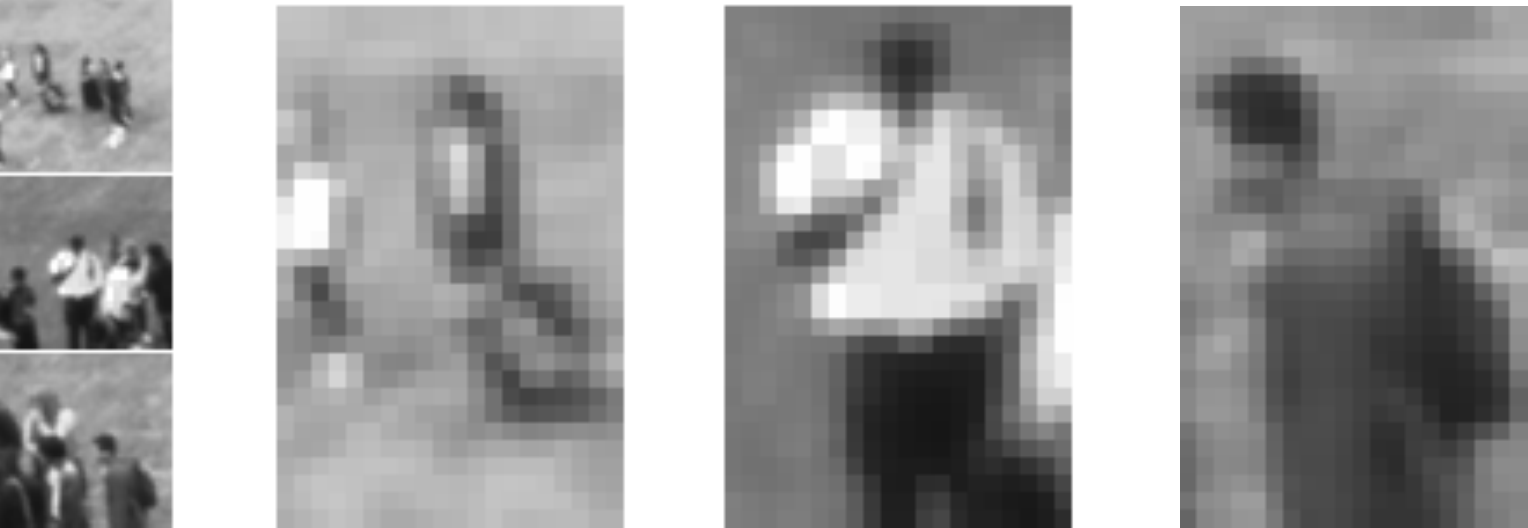}
      \caption{Left column, from top to bottom:  50 $\times$ 50 pixel patches from shots taken with increasing focal lengths ($f = 4mm$,  $f = 8mm$ and $f = 12mm$) from the same position. The following images, from left to right: interest areas from the three previous patches, of initial size 20 $\times$ 30 and zoomed for visualization purposes with no interpolation applied. Detailed features essential for scene analysis are not retrieved below a certain focal length.}
      \label{fig:lenspatches}
\end{figure*}

Nevertheless, we ought also take into account the actual aim of retrieving information related to the area of interest. For this purpose, we have acquired from the same position and with the same camera three shots using lenses with 4, 8 and 12 mm focals respectively. In the left column of Fig. \ref{fig:lenspatches} we present from top to bottom three 50 $\times$ 50 pixel patches from the shots taken with increasing focal lengths. The adjacent images from left to right show areas from these patches (of initial size 20 $\times$ 30 and zoomed for visualization purposes with no interpolation applied). In this case, the long focal lenses are required for retrieving with enough detail entities such as body parts, bags etc. which are essential for a wide range of tasks related to action understanding, monitoring, tracking and surveillance.

From the above illustrations, we may thus notice that for the purpose of analysis of a specific area of interest, a wide FOV is beneficial for accurate registration in a  camera network, whilst a narrow FOV is beneficial for retrieving details from the area of interest. These details are often homogeneous, and by lacking salient features the narrow FOV is not able to solve robustly or at all the relative pose problem. 

A calibration pattern visible from all views set on the area of interest can solve the relative pose problem. However, there are multiple applications where this solution is not practical. The area of interest may be far and thus quite large, or it may be inaccessible. During the analysis, the camera poses might change accidentally due to shocks, periodically due to vibrations, or by design (mobile observers); all these scenarios require frequent relative pose estimation updates. In the following paragraphs, we recall briefly some works that are relevant for the problem of multiple view detailed analysis \emph{and} relative pose estimation, highlighting their respective benefits and shortcomings for this scenario.

\subsection{Zooming cameras}
One possible solution is to deploy a network of cameras which use motorized zoom lenses. Each camera will switch from a wide FOV used for the relative pose estimation to a narrow FOV used for analysis. One major consequence is that the zooming process modifies the intrinsic and distortion parameters of the cameras. These parameters have to be re-estimated in order to perform either epipolar search or 3D-image plane correspondences, which are systematically used in camera networks with overlapping views. 

Various solutions for zooming recalibration based on the scene have been proposed; these solutions are often denoted as self-calibration methods. For optical center and distortion parameter estimation, common strategies make use of straight lines in the scene \citep{Ahmed05, Wang09} or interest point correspondences \citep{Kukelova11}. For the rest of the intrinsic parameters, these methods rely on matching salient features, usually interest points, of a rigid scene either between frames (see for example \cite{Sturm97} for proposing pre-calibration to model the interdependence of intrinsic parameters, or \cite{lourakis}), or with respect to an existing 3D scene model \citep{josephson}. Furthermore, some works investigate the self-calibration of a stereo system (intrinsic and extrinsic parameters), but they consider typically a subset of the intrinsic parameters. For example, \cite{Dang09} propose a scene based estimation method for the focal lengths and the relative rotation for a stereo system; however, the system being tested in a real scenario has fixed focal lengths, and the method is only shown to be robust to noisy initializations of the focal length values. Actual focal length variations would have an impact on other assumptions (i.e. fixed principal point or straightforward removal of distortion effects). Generally, the self-calibration scenarios make some simplifying assumptions (i.e. fixing the principal point, ignoring focus effects) that are inaccurate and thus have a detrimental effect on the projection function accuracy. 

\subsection{Pan-Tilt-Zoom (PTZ) cameras}
PTZ cameras have a built-in zooming function and are specifically designed for live monitoring. However, tasks such as surveillance or auto tracking  do not require necessarily accurate self-calibration. In the area of PTZ camera network calibration, the work introduced in \cite{sinha04} builds high resolution panoramas of the scene, which are then used for extrinsic calibration. This solution has been shown to be effective for static environments rich in salient features. In terms of tracking dynamic targets, \cite{Bimbo10} propose a wide FOV (master) to narrow FOV (slave) registration process, which is again limited by the presence of static landmarks. Another work, investigating this time the coupling between a static wide FOV camera and an active camera, has been presented in \cite{Horaud06}, with an application to gaze control. Also in order to cope with and exploit the presence of moving objects, some approaches actually rely on aligning trajectories for relative pose estimation, especially in the case of wide baseline stereo~\citep{Puwein12}; however, these solutions cannot be applied to calibration in scenarios such as crowded scenes when calibration itself is a prerequisite of successful tracking. 

\subsection{An argument for hybrid systems}
The strategies recalled up to this point propose interesting solutions for self-calibration and camera registration in the presence of a sufficient number of salient features, but they are not applicable for a camera view if the lens is zoomed on a \emph{homogeneous} and/or \emph{dynamic} scene. Although these scenarios are less common, examples of possible applications abound in the study of crowds and of different types of flows encountered in natural phenomena. The underlying idea for the solution we propose is about transferring pose information in a scene-independent manner to the zoomed camera from a secondary camera able to infer its pose. This leads to a straightforward minimal solution based on a rigid stereo rig featuring two cameras, one with a small FOV used for analysis, and one with a large FOV used for registration within a network of such rigs. 

Surprisingly, this solution has not been applied to the analysis of homogeneous scenes. Even considering a broader range of applications, the use of hybrid stereo systems featuring large and narrow FOV is limited. In robotics, the use of a hybrid setup has been illustrated recently by \cite{Eynard12} who employ a fisheye and a perspective camera on a UAV. The authors show how the richer information from both views may be used conveniently for a sequential estimation of the unknowns in the following order: attitude, altitude, then motion. Another application where a hybrid system has been used is the recent STEREO solar observation mission \citep{eyles2009heliospheric, Brown09}. Each of the two STEREO spacecraft features two heliospheric imagers, HI1 and HI2, with FOV of 20$^\circ$ and 70$^\circ$ respectively. The imagers do have a common FOV, but this property is used only for photometric cross-calibration, since both imagers are able to estimate their pose accurately with respect to a star catalogue. The main interest of using this setup is the ability to study the propagation of coronal mass ejections along the whole Sun-Earth line, with an increased resolution close to the Sun.

With respect to the previous works, the solution based on a hybrid stereo system has clear benefits for the analysis of dynamic homogeneous scenes. As long as the study within a specific region is applicable, the advantage of the fixed calibration parameters is twofold. We do not have to adopt any simplifying assumptions about the variations of intrinsic parameters, and the calibration precision will be maintained at the optimal level provided by state of the art calibration algorithms. Secondly, the extrinsic parameters of each stereo rig can be estimated independently of the scene. Unlike in the scenario of a zooming camera, the availability at each instant of an accurately registered pair of a high resolution image and of a panoramic image of the scene allows for accurate pose re-estimation between rigs as often as necessary, overcoming the effect of movement and vibrations. 

\section{Background on scene based pose estimation}
\label{sec:background}
\subsection{The projection model}
\label{subsec:model}
In the following, we will briefly recall the pinhole camera and optical distortion models that we employ. A point in 3D space $\mathbf{X} = [\mathrm{X} \; \mathrm{Y} \; \mathrm{Z}]^\mathrm{T}$ projects within the image space into a pixel $\mathbf{x} = [x \; y]^\mathrm{T}$ according to:
\begin{equation}
\left ( 
  \begin{array}{c}
    \mathbf{x} \\
    1
  \end{array}
\right )
  =\lambda \mathbf{K} \Big[ \mathbf{R} \; |\; \mathbf{-RC} \Big ]
\left ( 
  \begin{array}{c}
    \mathbf{X} \\
    1
  \end{array}
\right )
\end{equation}
with $\lambda$ being an undetermined scale factor, $\mathbf{R}$ the orientation of the camera and $\mathbf{C}$ the location of its optical center in world coordinates (we also note $\mathbf{t}=-\mathbf{R}\mathbf{C}$), and $\mathbf{K}$ the intrinsic parameters:
\begin{equation}
\mathbf{K} = 
\left [ 
  \begin{array}{ccc}
    f_x & s & c_x \\
    0 & f_y & c_y \\
    0 & 0 & 1 
  \end{array}
\right ]
\end{equation}
Above, $f_x$ and $f_y$ are the focal lengths, $[c_x\; c_y]^\mathrm{T}$ represents the principal point, and the skew parameter $s$ is considered 0. 

In order to switch to different coordinate frames, we rely on elements of $\mathrm{SE}(3)$, the group of rigid body transformations in $\mathbb{R}^3$. A transformation matrix $\mathbf{E}$ takes the form:

\begin{equation}
\mathbf{E} = 
\left [ 
  \begin{array}{cc}
    \mathbf{R} & \mathbf{t} \\
    0 & 1 
  \end{array}
\right ]
\end{equation}
Element multiplication amounts to transitive chaining coordinate frame transformations: $\mathbf{E}^{CA} = \mathbf{E}^{CB}\mathbf{E}^{BA}$ would transfer a 3D point in homogeneous coordinates from reference system $A$ to reference system $C$, based on both $A$ and $C$ relative relations to reference system $B$.

In order to account for radial distortion, the extension of the pinhole model assumes that if the 3D point $\mathbf{X}$ is projected to $[\tilde{x} \; \tilde{y} \; 1]^\mathrm{T}$ under the initial assumptions, then $\mathbf{X}$ would be actually imaged to the distorted location $[x_d \; y_d]^\mathrm{T}$ :
\begin{equation}
\left ( 
  \begin{array}{c}
    x_d \\
    y_d
  \end{array}
\right )
  =(1+\sum_{i=1}^3 \kappa_i \tilde{r}^{2i})
\left ( 
  \begin{array}{c}
    \tilde{x} \\
    \tilde{y}
  \end{array}
\right )
\end{equation}
where $\tilde{r} = \left(\tilde{x}^2 + \tilde{y}^2 \right)^{1/2}$. Thus, $(f_x, f_y, c_x, c_y, \kappa_1, \kappa_2, \kappa_3)$ is in most scenarios the suitable parameter set for a full intrinsic calibration.

\subsection{Epipolar geometry}
\label{subsec:epigeom}

One tool that we will employ in the following sections is the epipolar constraint, which is a direct implication of the projective geometry between two views. It is worth noting that this constraint is independent of the scene structure, depending exclusively on the intrinsic parameters and the relative pose - as long as the the salient features of the scene are static, or as long as the cameras are accurately synchronized.

Considering two projections $\mathbf{x}_1$ and $\mathbf{x}_2$ of the same point $\mathbf{X}$ in cameras $C_{1}$ and $C_{2}$, the epipolar constraint defines the relationship between the projections as:
\begin{equation} 
\mathbf{x}_2^\mathrm{T} \mathbf{F} \mathbf{x}_1 = 0
\end{equation} 

In the above, $\mathbf{F}$ is known as the fundamental matrix \citep{Faugeras92}, which depends explicitly on the calibration parameters in the following way:
\begin{equation} 
\mathbf{F} = \mathbf{K}_{2}^{-\mathrm{T}}\mathbf{t}_{\times}\mathbf{R}\mathbf{K}_{2}^{-1}
\end{equation} 
where $\mathbf{t}_{\times}$ is the skew-symmetric matrix associated to $\mathbf{t}$.

The main interest of the epipolar constraint is that it does not make any assumptions about the 3D structure of the scene. Thus, compared to other optimisation algorithms that are commonly employed to estimate the relative pose, the determination of  $(\mathbf{R}, \mathbf{t})$ using the epipolar geometry provides a practical minimal parametrization and does not require an initialization. However, the result may be used for the initialization of more complex optimisations, such as the bundle adjustment procedure, briefly recalled in the following section. 

\subsection{Bundle adjustment}

Assuming a zero-mean Gaussian distribution of the corner detection errors, bundle adjustment \citep{Triggs99, Hartley2004} is the Maximum Likelihood Estimator for the joint estimation problem of relative camera poses and of observed 3D point locations. The bundle adjustment procedure will minimize the following reprojection error:
\begin{equation}
\min_{\hat{P}^{i}, \hat{\mathbf{X}}_{j}} \sum_{i,j} d\bigg (\hat{P}^{i}(\hat{\mathbf{X}}_{j}), \mathbf{x}_{j}^{i}\bigg )^2
\end{equation}
In the error function above, $\hat{\mathbf{X}}_{j}$ is the location hypothesis for a point observed by the $i^{th}$ camera. The projection function $\hat{P}^{i}$ related to the pinhole model (accounting for radial distortion too) depends on the $i^{th}$ camera pose; we consider that the intrinsic parameters are known and are not part of the optimization problem. The bundle adjustment will thus minimize jointly for all the possible camera-point pairs $(i,j)$ the distance between the reprojection $\hat{P}^{i}(\hat{\mathbf{X}}_{j})$ and the actual measurement $\mathbf{x}_{j}^{i}$. Solving this optimization problem is studied in depth in the literature, and it generally boils down to exploiting the sparsity of its Hessian matrix and to employing an adapted non-linear least squares algorithm such as Levenberg-Marquardt \citep{lour09}.

Although bundle adjustment seems like an ideal solution for multiple view pose estimation, it does have some well-known shortcomings that we will briefly discuss in connection with our specific aim. One common criticism is related to the computational requirements, but this issue is more prevalent in large scale robotics applications, especially if there are real-time constraints to take into account. For a relatively small camera network, the size of the problem is reasonable even for frequent updates. Another important aspect is related to the initialization, which has to be relatively accurate in order to allow the problem to converge to the correct solution. In order to cope with this, we will rely on an initialization based on the epipolar constraint discussed in Section \ref{subsec:epigeom}, but other options are possible too (see for example \citealp{Nister04}, or \citealp{Kneip11} if 3D information about some scene features is available). Finally, some practical aspects are equally relevant. Given the  high number of parameters which are usually involved, constraining the relative pose variables is more effective if the adjacent camera views for the large FOV cameras are close enough in order to allow for a significant FOV overlap. Stability is also improved if the corner correspondences are spread onto the common field of view. 

\section{The proposed algorithm}
\label{sec:algo}

\subsection{Outline}

Let us consider a network of $N$ hybrid stereo rigs, the $i^{th}$ rig featuring a small FOV camera $C_{i}^s$ used for analysis, and a large FOV camera $C_{i}^l$ employed for pose estimation in a global frame. The aim of the following procedure is to align accurately the cameras $\{C_{1}^s,C_{2}^s,\ldots, C_{N}^s \}$. We assume that for each rig, the cameras $C_{i}^s$ and $C_{i}^l$ have been calibrated. In the following, we will denote by $\mathbf{E}_{i}^{sl}$ the transform that transfers a point from the large FOV camera to the analysis camera on the $i^{th}$ stereo rig. Also, $\mathbf{E}_{ji}^{l}$ and $\mathbf{E}_{ji}^{s}$ are transforms that transfer points from the $i^{th}$ rig to the $j^{th}$ between the large FOV cameras, and respectively between the analysis cameras.

The fact that the stereo rigs are passive allows for a precise intrinsic and extrinsic calibration which can be performed independently of the scene in a controlled environment. Thus the intrinsic parameters $\mathbf{K}_{i}^s$, $\mathbf{K}_{i}^l$ as well as the rigid transform $\mathbf{E}_{i}^{sl}$ that projects a 3D point from the pose estimation camera of the rig to the analysis camera are considered as known. 

For the next step, let us consider a pair of spatially adjacent rigs  $(i,j)$; in most scenarios, cameras are spread as much as possible, and thus it is necessary to consider adjacent pairs in order to obtain enough reliable interest point matches. Due to initialization requirements, we cannot apply bundle adjustment directly in order to estimate $\mathbf{E}_{ji}^{l}$ between the two large FOV cameras on the rigs. We perform SIFT detection and matching \citep{Lowe04}, and use the normalized 8-point algorithm \citep{Hartley97} with RANSAC \citep{Fischler81} for robustness to outliers. For the matching step, we employ two filtering strategies based on the uniqueness assumption (the ratio $\tau$ of the similarity scores for the top two candidates, \citep{Lowe04} and on married matching (both features are the top candidate for each other \citep{NisterNB04}). Then, we decompose the fundamental matrix \citep[chap. 9]{Hartley2004} and choose the correct solution based on the chirality constraint \citep{Hartley98}. Let us denote $\tilde{\mathbf{E}}_{ji}^{l}$ the rigid transformation estimated after this step. Using $\tilde{\mathbf{E}}_{ji}^{l}$, and based on the inlier set of matches that were validated during the RANSAC procedure, we build a set of 3D points $\tilde{\mathbf{X}}_{ji}$ by linear triangulation \citep{HartleyS97}. 

At this point, we can employ a bundle adjustment procedure using $\tilde{\mathbf{E}}_{ji}^{l}$ and $\tilde{\mathbf{X}}_{ji}$ as initial estimates, and we obtain a refined relative pose estimation $\hat{\mathbf{E}}_{ji}^{l}$ for the large FOV cameras in the pair of rigs $(i,j)$. Ideally, a bundle adjustment involving more than a pair of rigs should be performed afterwards whenever possible; however, in a typical setting, cameras are spaced as much as possible around a scene, the limit being imposed by common FOV considerations and the performance of the interest point matching procedure. Thus, we may assume that in most situations non-adjacent rigs will have difficulties for the matching procedure, and will have matches corresponding to disjoint sets of 3D points, which effectively yields the bundle adjustment problems independent. A particular setting is that of a scene surrounded in a full circle by rigs, and in this case a full bundle adjustment may be beneficial.

Having the bundle adjustment estimations, it is trivial to express the $C_{i}^l$ poses in a common reference system; in the following, we set this reference system as depicted by the position and orientation of $C_{1}^l$. Let $\hat{\mathbf{E}}_{i}^{l}$ be the rigid transform that links $C_{1}^l$ to $C_{i}^l$. For any two rigs $(i,j)$, we can now use the extrinsic calibrations $\mathbf{E}_{i}^{sl}$, $\mathbf{E}_{j}^{sl}$ and the global allignment of the large FOV cameras in order to infer the global allignment of the analysis cameras in the same reference system, as well as their relative pose:
\begin{equation}
\mathbf{E}_{i}^{s} = \mathbf{E}_{i}^{sl} \hat{\mathbf{E}}_{i}^{l}; 
\mathbf{E}_{j}^{s} = \mathbf{E}_{j}^{sl} \hat{\mathbf{E}}_{j}^{l};
\end{equation}
\begin{equation}
\mathbf{E}_{ji}^{s} = \mathbf{E}_{j}^{sl}\hat{\mathbf{E}}_{j}^{l} \left( \mathbf{E}_{i}^{sl} \hat{\mathbf{E}}_{i}^{l} \right )^{-1}
\end{equation}

\subsection{Enforcing a common scale}
Bundle adjustment can estimate accurately the relative pose up to an unknown scale factor. This limitation applies to the $\hat{\mathbf{E}}_{ji}^{l}$ estimates only; the values $\mathbf{E}_{i}^{sl}$ that specify the baseline for cameras on the same rig are not concerned as long as a known size calibration pattern is used for stereo extrinsic calibration. Since the different bundle adjustment procedures depicted in the following paragraphs are typically independent, we have to enforce a common scale factor among all optimizations using additional information. Depending on the application, it is easier to adopt one of the following strategies:
\begin {itemize}
\item for a small sized scene, add a known size object in the common FOV of $C_{i}^l$  and $C_{j}^l$; we thus use $\tilde{\mathbf{X}}_{ji}$ in order to impose a metric scale to the reconstruction
\item for a large scene, either use a similar approach as for the previous setting, or if it is not applicable measure the distance between $C_{i}^l$  and $C_{j}^l$ (using for example a laser rangefinder), thus using $\tilde{\mathbf{t}}_{ji}$ in order to impose a metric scale to the reconstruction.
\end {itemize}

The main steps of the algorithm are synthetically recalled below.

\begin{algo}
\begin{small}
relative pose estimation for homogeneous scene analysis
\\
\\
\noindent\textbf{Objective}: given a network of $N$ hybrid stereo rigs $(C_{i}^s;C_{i}^l)$, estimate for any pair $(i,j)$ the rigid transform $\mathbf{E}_{ji}^{s}$ that projects a 3D point from the analysis camera of the $i^{th}$ rig to the analysis camera of the $j^{th}$ rig.

\noindent\textbf{Steps}: 
\begin{enumerate}[label=(\roman{*}), ref=(\roman{*}), leftmargin=*]
\item For all rigs, estimation of  $\mathbf{K}_{i}^{s}$, $\mathbf{K}_{i}^{l}$ 
\item For all rigs, estimation of  $\mathbf{E}_{i}^{sl}$

\item For each pair of adjacent rigs $(i,j)$:
\begin{enumerate}%[label=\arabic*, leftmargin=*, noitemsep]
\item estimate $\mathbf{F}_{ji}^{l}$
\item estimate $\tilde{\mathbf{R}}_{ji}^{l}$, $\tilde{\mathbf{t}}_{ji}^{l}$, $\tilde{\mathbf{E}}_{ji}^{l}$
\item use $\tilde{\mathbf{E}}_{ji}^{l}$ to triangulate matches and obtain $\tilde{\mathbf{X}}_{ji}$
\item enforce a metric scale, either through $\tilde{\mathbf{t}}_{ji}^{l}$ or through $\tilde{\mathbf{X}}_{ji}$
\item Apply bundle adjustment using  $\tilde{\mathbf{E}}_{ji}^{l}$, $\tilde{\mathbf{X}}_{ji}$ for initialization, and compute $\hat{\mathbf{E}}_{ji}^{l}$, $\hat{\mathbf{X}}_{ji}$ 
\end{enumerate}
\item register all $\hat{\mathbf{E}}_{i}^{l}$ in the reference frame of $C_{1}^l$
\item if applicable, reapply bundle adjustment for more than two adjacent rigs at a time, using previously computed values for initialization
\end{enumerate}
\noindent\textbf{Result}: for given $(i,j)$, compute $$\mathbf{E}_{ji}^{s} = \mathbf{E}_{j}^{sl}\hat{\mathbf{E}}_{j}^{l} \left( \mathbf{E}_{i}^{sl} \hat{\mathbf{E}}_{i}^{l} \right )^{-1}$$

\end{small}
\label{algo1}
\end{algo}

\section{Experimental results}
\label{sec:results}

\subsection{A small scale scenario}

\begin{figure*}[htbp]
 \begin{center}

 \begin{minipage}{0.94\textwidth}\centering
   \includegraphics[width=\textwidth]{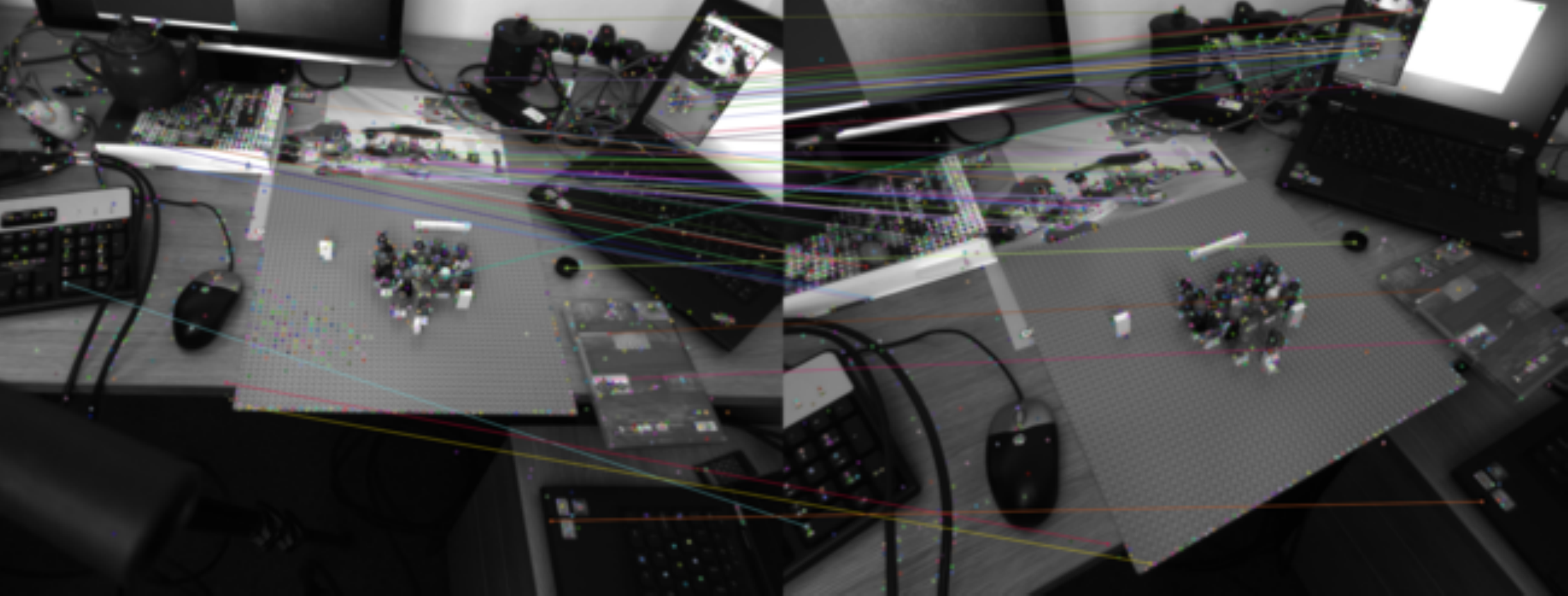}\\
   {\small a) }
  \end{minipage}
 \\
 \begin{minipage}{0.94\textwidth}\centering
  \includegraphics[width=\textwidth]{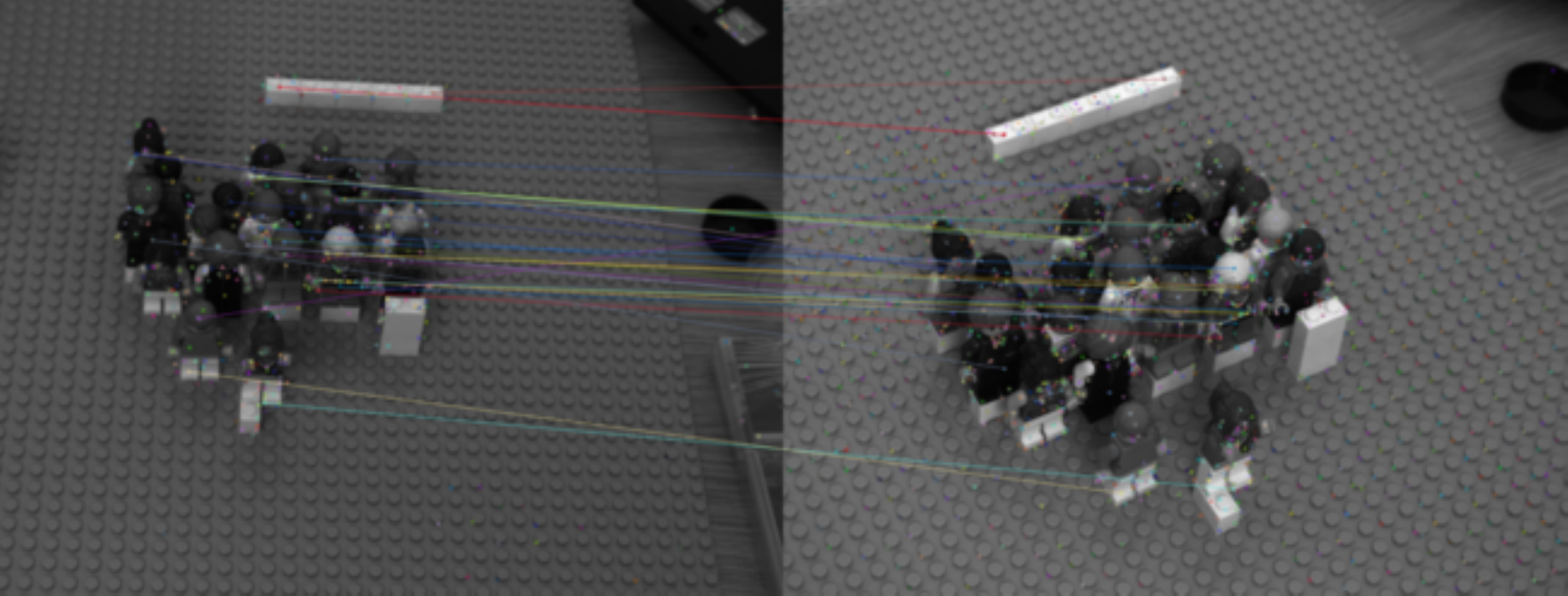}\\
  {\small b) }
 \end{minipage}
 \end{center}
 \caption{\small A set of images used for pose estimation in a simple indoor environment; the images in a) correspond to $C_{1}^l$ and $C_{2}^l$, and the images in b) show the images captured by $C_{1}^s$ and $C_{2}^s$. Both pairs of images have been matched using SIFT; the first set of matches are necessary for the algorithm, whilst the second set is used as ground truth in order to verify the result of the algorithm.}
 \label{fig:multiLEGO}
\end{figure*}

We have created a simple example in an indoor environment, using LEGO figurines placed closely in the middle of a homogeneous surface. We have used two hybrid stereo rigs and taken a snapshot of the figurines and surrounding environment. The resulting images are presented in Figure \ref{fig:multiLEGO}: the upper and lower rows show the views from the large FOV ($C_{1}^l$,$C_{2}^l$) and small FOV ($C_{1}^s$,$C_{2}^s$) cameras respectively. We have also highlighted the results of the matching procedures; the first matching set ($\tau = 0.4$ for uniqueness) is required for Step (iii) of the algorithm, while the second (a more permissive value $\tau = 0.75$ has been set in order to have enough matches) is not used in the algorithm - as the scene is supposed to be poor in salient features - but it is used as ground truth for validating the result of the algorithm. We apply the steps highlighted in the previous Section in order to compute $\mathbf{E}_{21}^{s}$: estimation of $\mathbf{E}_{21}^{l}$ using SIFT matching followed by decomposition of $\mathbf{F}_{21}^{l}$ and bundle adjustment, then exploitation of $\mathbf{E}_{1}^{sl}$ and $\mathbf{E}_{2}^{sl}$ provided by stereo calibration, and also the setup of the right scale by using an object of known size (the long brick of length 79.8 mm).

In order to estimate numerically the quality of the rigid transform $\mathbf{E}_{21}^{s}$ obtained, we have exploited the matches that we were able to determine directly between the small FOV cameras in this example. In homogeneous scenes, interest points may be completely absent, or the scarcity of matches may have a detrimental effect on the stability of the estimation of $\mathbf{F}$. Therefore, in our example we set up a base bundle adjustment problem between $C_{1}^s$ and $C_{2}^s$ where we initialize the system by the decomposition of $\mathbf{F}_{21}^{s}$. Alternatively, we use the rigid transform $\mathbf{E}_{21}^{s}$ as initialization for the triangulation of matches and for the BA procedure. The resulting solutions and mean reprojection errors for these two scenarios are presented in Table \ref{numbersLEGO}. 

As we notice, the two optimization problems converge towards the same solution, but $\mathbf{E}_{21}^{s}$ brings the optimization much closer to the objective in terms of mean reprojection error. This result is interesting for a number of reasons. Firstly, even though we do not have a case of optimization stuck in a local minimum due to the worse initialization, this is a good example of coarse to fine resolution of the relative pose estimation. This approach is helpful for robotics applications in case of unstable optimizations (few matches in the small FOV cameras), and also interesting for the computation gain due to a faster convergence of BA (25 iterations with initial subpixel mean reprojection error compared to 37 iterations). Secondly, and most importantly, this example shows that in cases where we can not compute $\mathbf{E}_{21}^{s}$ directly due to the complete absence of salient features, we are able using this algorithm to infer the unknown rigid transform from the adjacent large FOV cameras with a high level of accuracy.

\begin{table*}[htbp]
\caption{ Relative poses between $C_{1}^s$ and $C_{2}^s$. The Euler angles are expressed in degrees, and the mean reprojection errors in pixels. Tilde values represent estimations prior to the bundle adjustment procedure, and hat values denote estimations refined by the bundle adjustment. The difference between the two rows consists in the initialization of the bundle adjustment; in the first case we use the SIFT matches depicted in Figure \ref{fig:multiLEGO}b), whilst in the second case we use the result of our algorithm. 
}
\begin{adjustwidth}{-1in}{-1in}% adjust the L and R margins by 1 inch
\begin{center}
{\footnotesize
\begin{tabular}{c|c|c|c|c|c|c|c}
 $( \tilde{\bm{\psi}}; \tilde{\bm{\theta}}; \tilde{\bm{\phi}}) $  & $\tilde{\mathbf{C}}$ & $ \tilde{\bm{\epsilon}}$ & $( \hat{\bm{\psi}}; \hat{\bm{\theta}}; \hat{\bm{\phi}}) $  & $\hat{\mathbf{C}}$ & $ \hat{\bm{\epsilon}}$ & {\bfseries Iter. } & {\bfseries Observations }\\ 

\hline 
\hline
 $ \left (
      \begin{array}{c}
            24.13 \\
            21.04 \\
            10.67
      \end{array}
      \right ) $ & 
      $\left (
      \begin{array}{c}
            -0.89 \\
            -0.30 \\
             0.33
      \end{array}
      \right ) $ & 37.16 & $ \left (
      \begin{array}{c}
            23.95 \\
            21.13 \\
            3.74
      \end{array}
      \right ) $ & 
      $\left (
      \begin{array}{c}
            -0.79 \\
            -0.25 \\
             0.55
      \end{array}
      \right ) $ & 0.199 & 37 & Base solution \\ 
\hline 
$ \left (
      \begin{array}{c}
            23.85 \\
            16.42 \\
            3.53
      \end{array}
      \right ) $ & 
      $\left (
      \begin{array}{c}
            -0.77 \\
            -0.23 \\
            0.59
      \end{array}
      \right ) $ & 0.489 & $ \left (
      \begin{array}{c}
            23.95 \\
            21.13 \\
            3.74
      \end{array}
      \right ) $ & 
      $\left (
      \begin{array}{c}
            -0.79 \\
            -0.25 \\
             0.55
      \end{array}
      \right ) $ & 0.199 & 25 & Init. by  $\mathbf{E}_{21}^{s}$ \\
\hline
\end{tabular}
 } %for small
\end{center}
\end{adjustwidth}
\label{numbersLEGO}
\end{table*}

\subsection{Pose estimation for highly dense crowds}

\begin{figure}[htpb]
      \centering
      \includegraphics[width=0.47\textwidth]{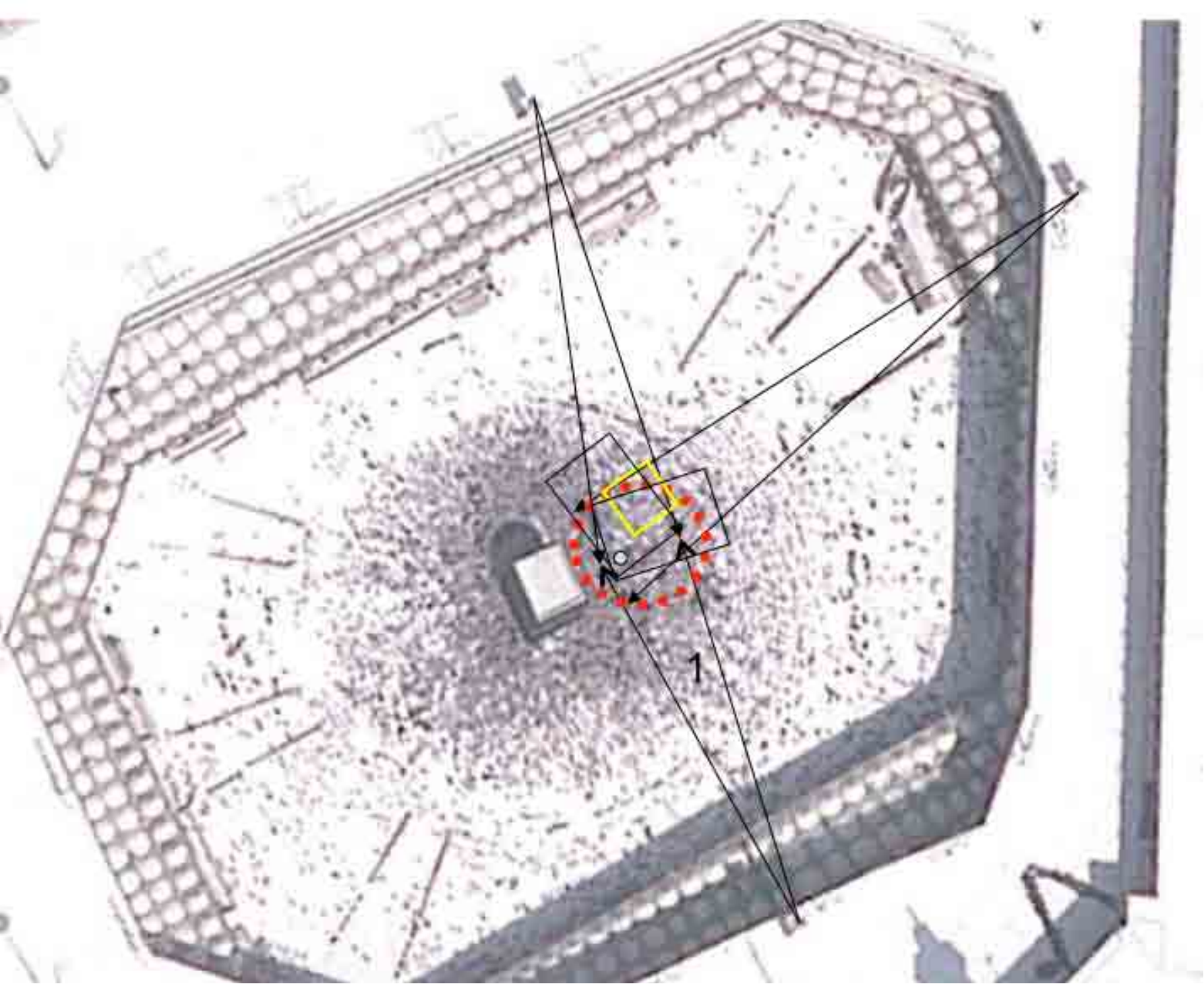}
      \caption{Birds eye view of the 'Mataf' area in the grand mosque at Makkah, and accessible locations for camera deployment; we have used the two upper locations (image obtained from Google Maps).}
      \label{fig:areaM}
\end{figure}

We have deployed two hybrid stereo rigs at the grand mosque in Makkah during very congested times of the Hajj period, in October 2012 (Figure \ref{fig:areaM}). The access constraints to the site impose a large perspective change between the two points of observation. As a result, neither SIFT nor ASIFT \citep{asift} algorithms were capable to provide any correct matches which are required as inputs for the algorithm we propose. Consequently, we had to rely on manual matching of salient structures in order to bootstrap the algorithm. 

\begin{figure*}[thbp]
 \begin{center}

 \begin{minipage}{0.47\textwidth}\centering
   \includegraphics[width=\textwidth]{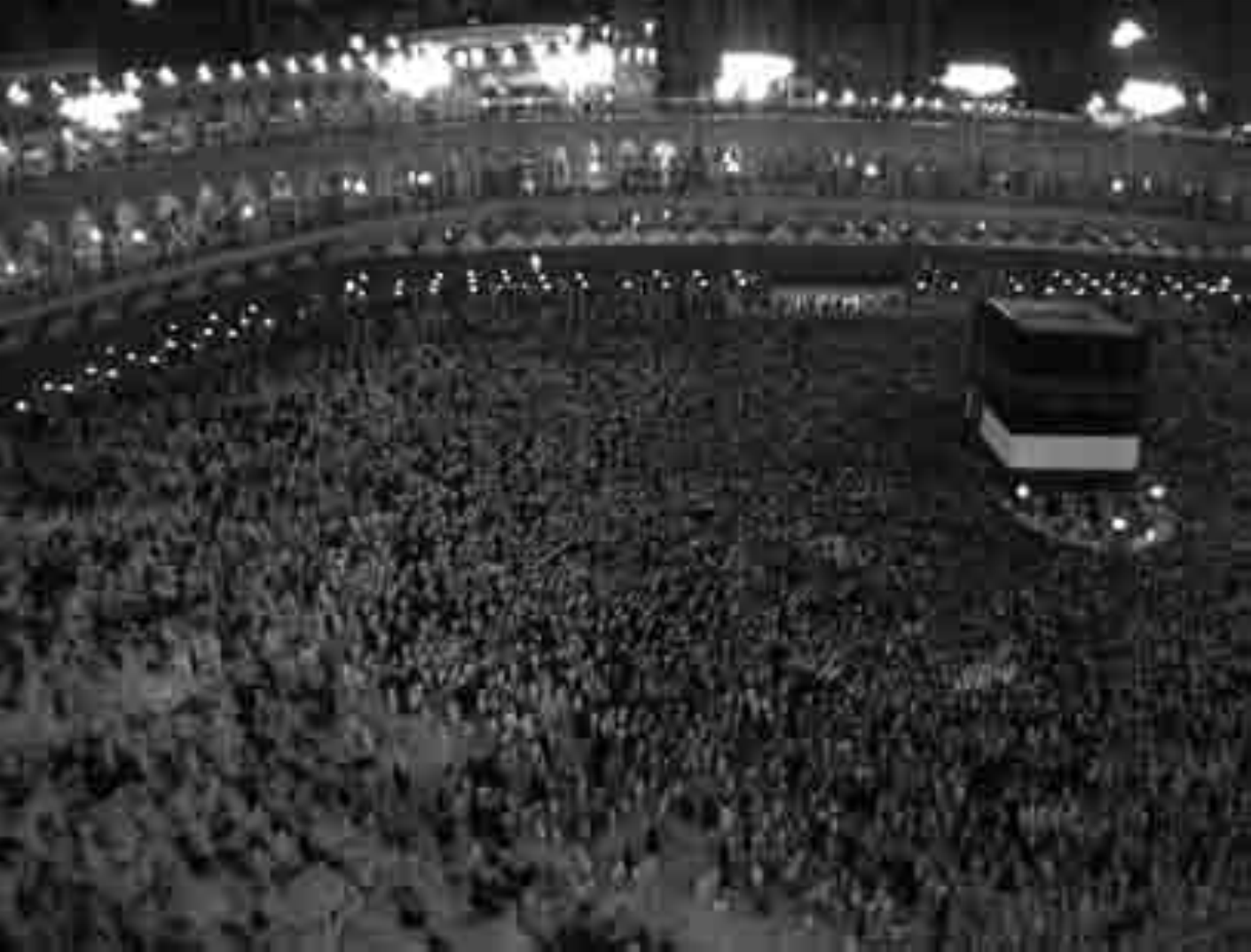}\\
   {\small a) }
  \end{minipage}
 \begin{minipage}{0.47\textwidth}\centering
 \includegraphics[width=\textwidth]{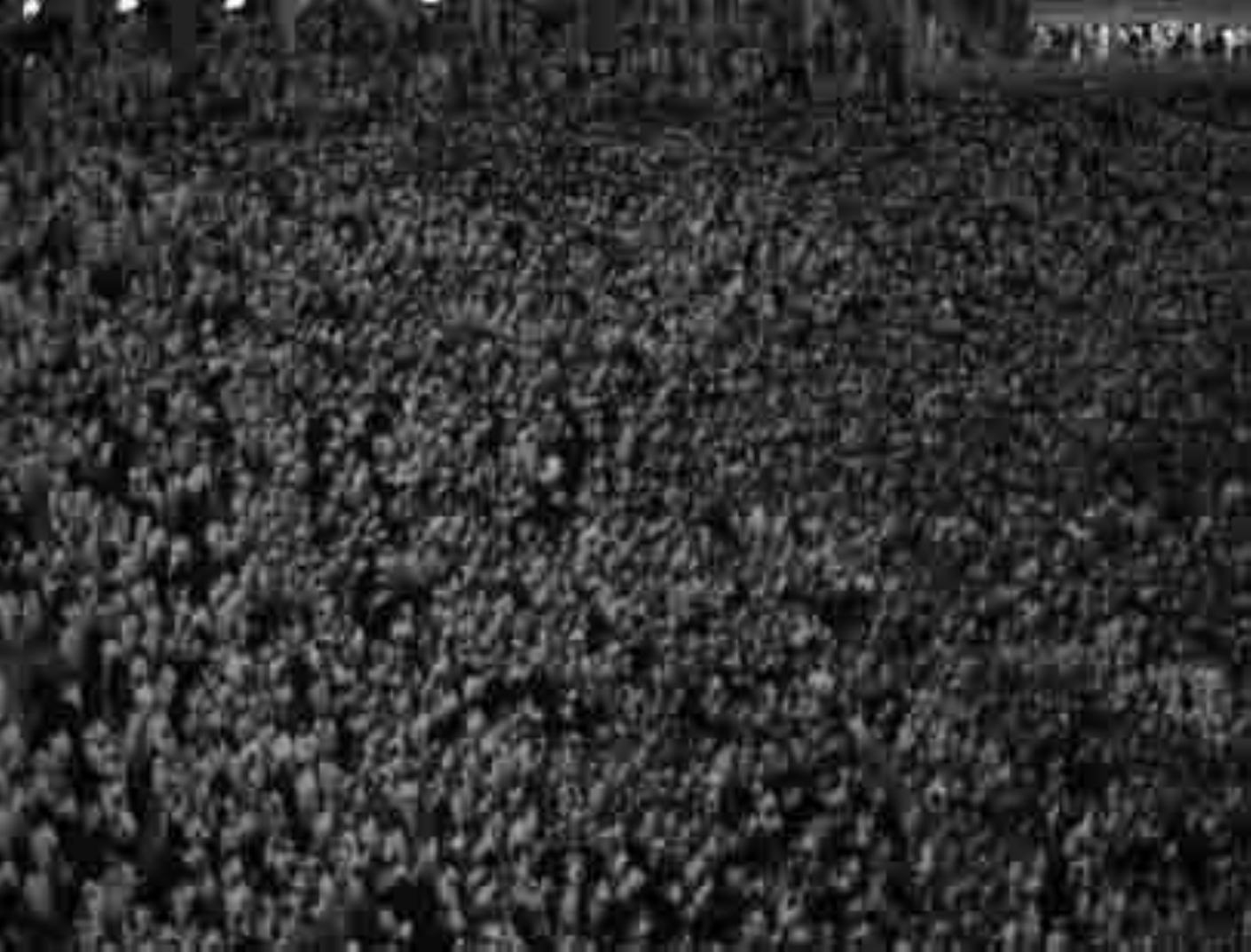}\\
  {\small b) }
 \end{minipage}
 \\
 \begin{minipage}{0.47\textwidth}\centering
  \includegraphics[width=\textwidth]{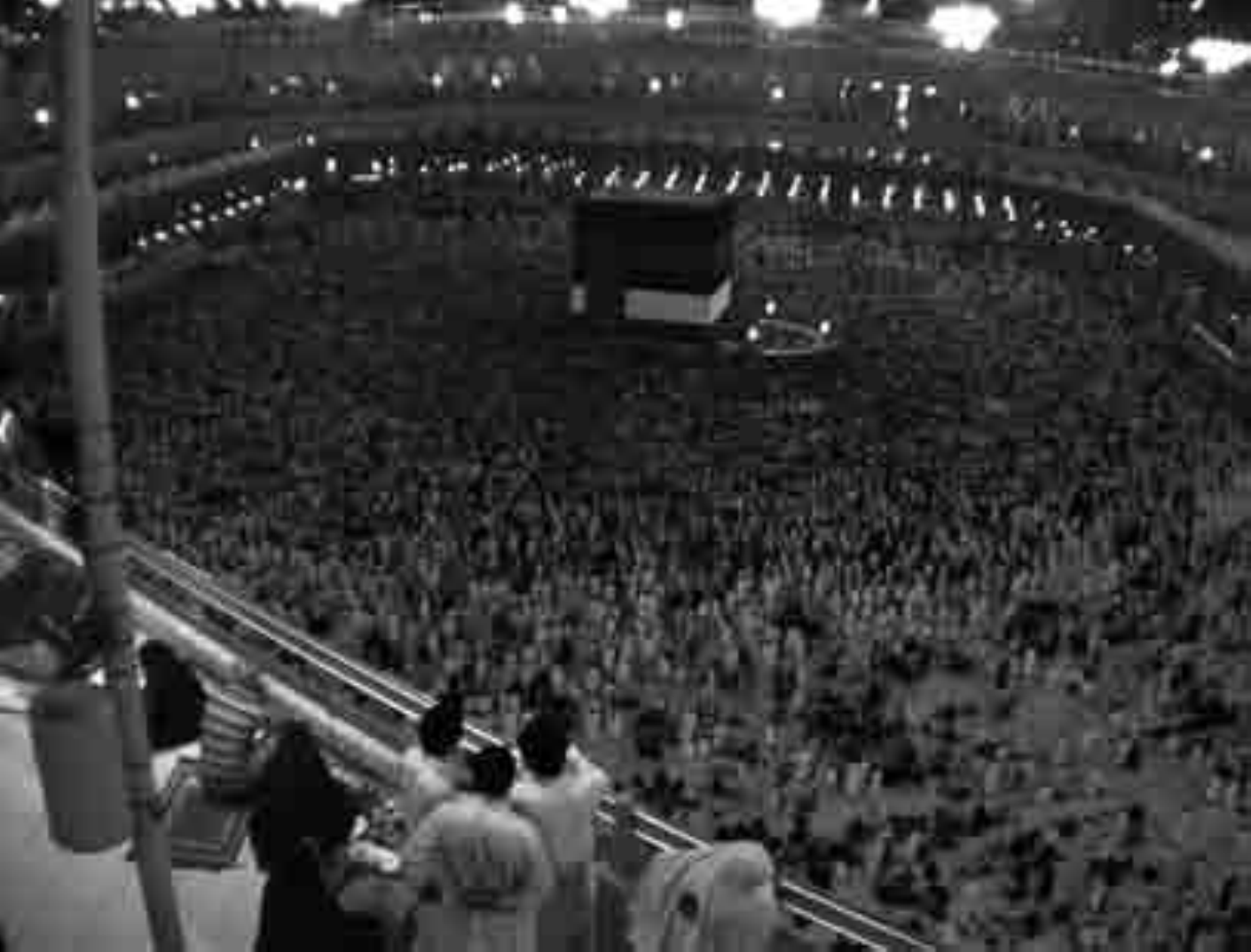}\\
  {\small c) }
 \end{minipage}
 \begin{minipage}{0.47\textwidth}\centering
 \includegraphics[width=\textwidth]{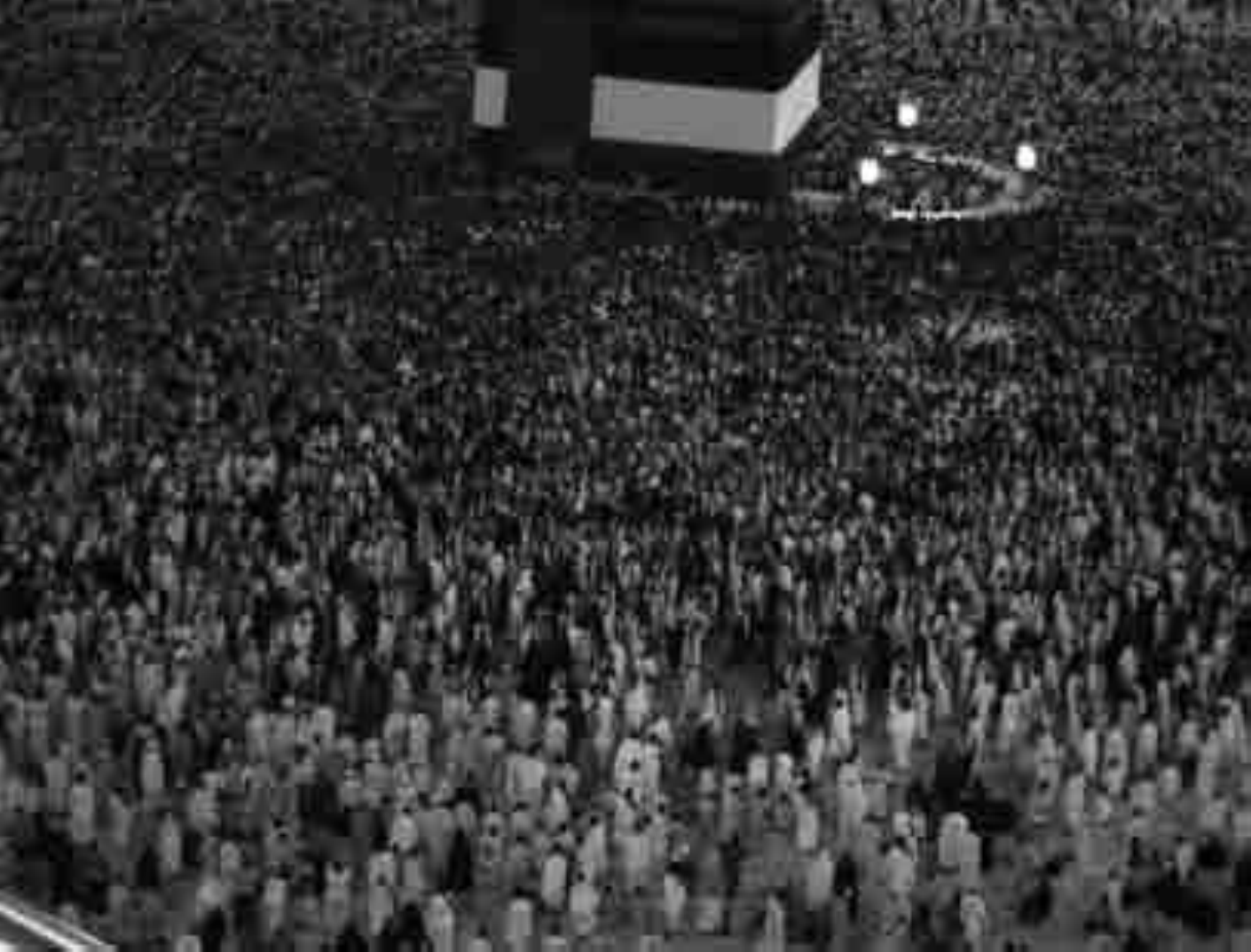}\\
  {\small d) }
 \end{minipage} 
 \\
 \begin{minipage}{0.19\textwidth}\centering
  \includegraphics[width=\textwidth]{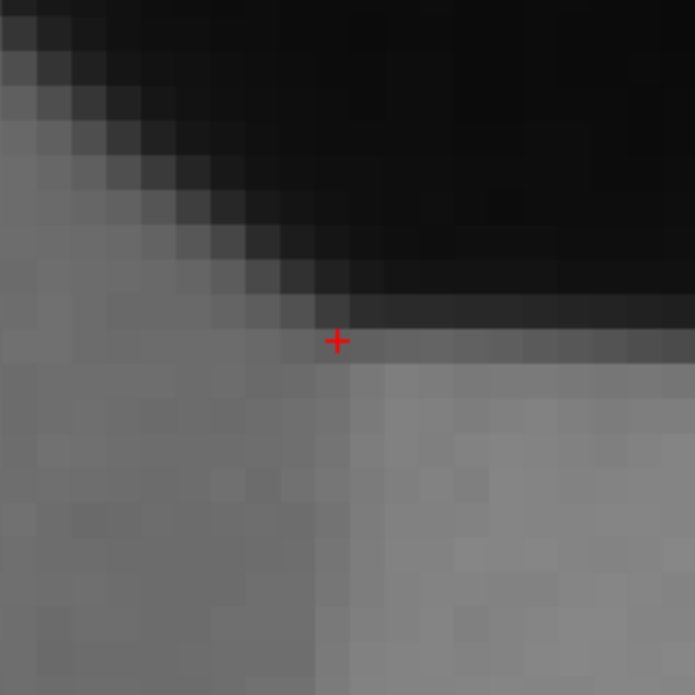}\\ 
  \includegraphics[width=\textwidth]{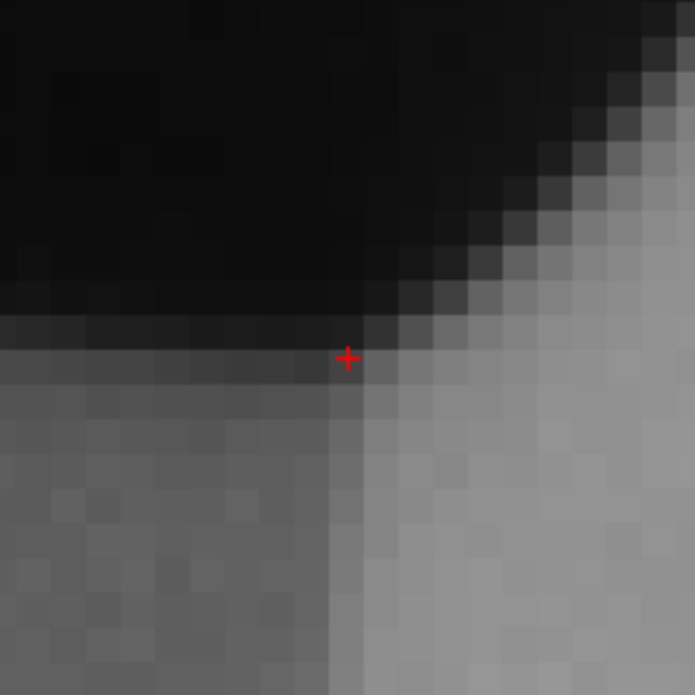}
  {\small e) }
 \end{minipage} 
 \begin{minipage}{0.38\textwidth}\centering
   \includegraphics[width=\textwidth]{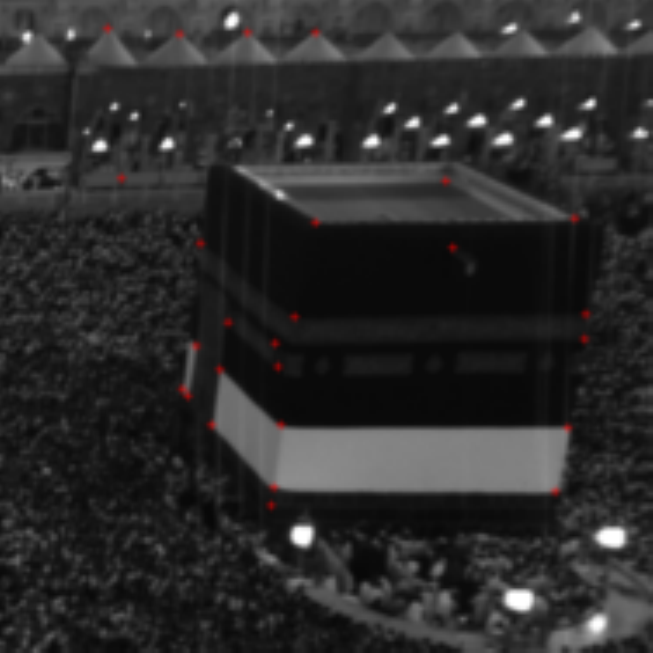}\\
   {\small f) }
  \end{minipage}
 \begin{minipage}{0.38\textwidth}\centering
 \includegraphics[width=\textwidth]{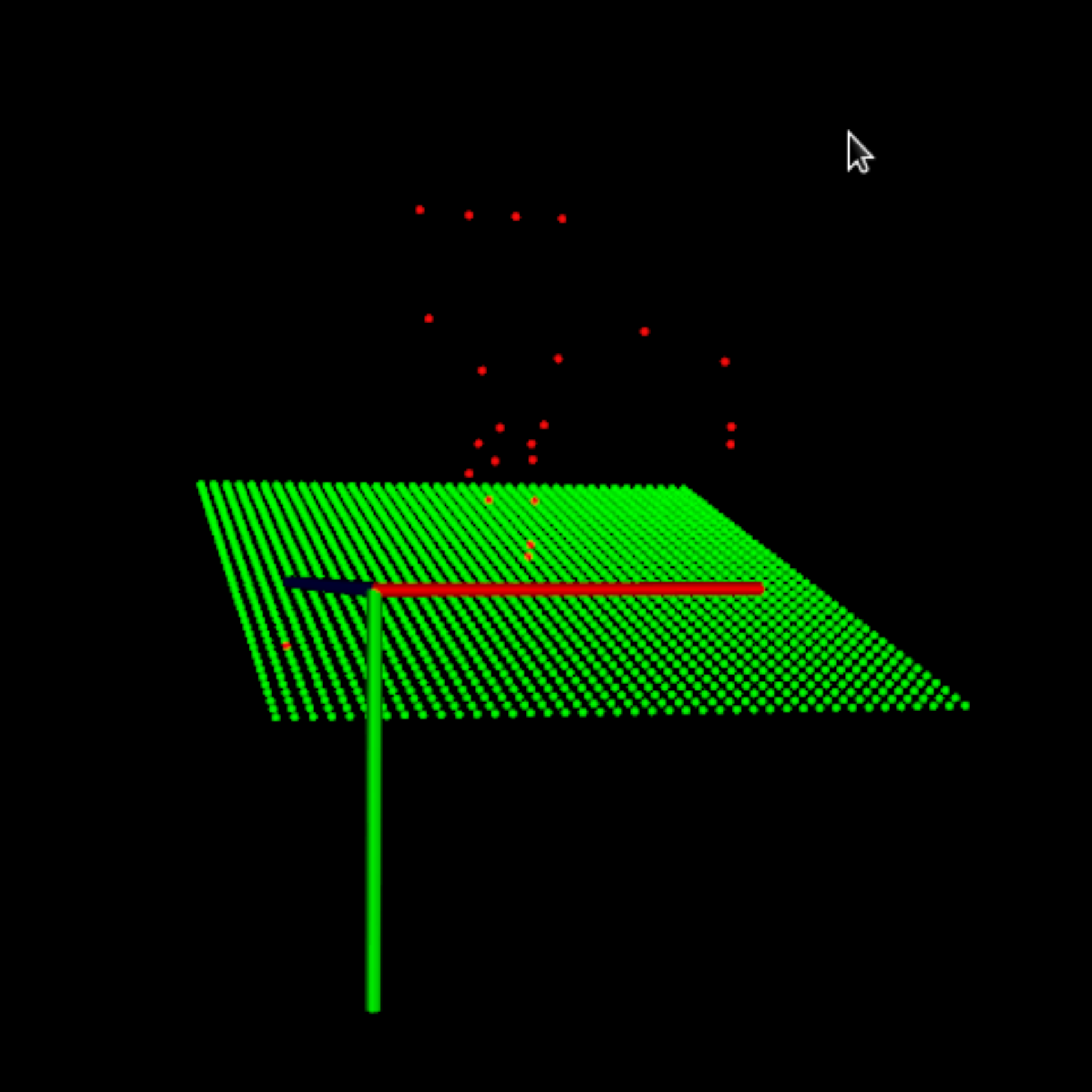}\\
  {\small g) }
 \end{minipage}
 
 \end{center}
 \caption{\small A set of images used for pose estimation; the images in a) and b) correspond to $C_{1}^l$ and $C_{1}^s$, and c) and d) correspond to $C_{2}^l$ and $C_{2}^s$ respectively. An example of user specified correspondences is illustrated in e). In f) we present the interest points used in the central region of one of the images, and in g) the inferred camera orientation (RGB axis for XYZ), with the approximate ground plane highlighted in green, for easier visualization.}
 \label{fig:multiM}
\end{figure*}

In Figure \ref{fig:multiM} we present the data our algorithm processed and registered; images in a) and b) correspond to $C_{1}^l$ and $C_{1}^s$, and c) and d) correspond to $C_{2}^l$ and $C_{2}^s$ respectively. The large FOV cameras contain enough common salient features, although the perspective variation does not allow for automated matching, and human intervention is necessary. Figure \ref{fig:multiM}e) presents such a user specified correspondence; in total we have used 34 user specified correspondences, of which 26 have been considered inliers for the fundamental matrix evaluation. For visualization purposes, Figures \ref{fig:multiM}f) and \ref{fig:multiM}g) present the central structure with the manually matched features, and the 3D structure of the scene with the camera axis aligned and an approximate representation of the ground plane. 

The numerical results of the algorithm for this setting are presented in Table \ref{numbers}; the relative rotations are expressed in degrees using Euler angles, the relative center position is expressed as a unit $\mathbb{R}^3$ vector, and mean reprojection errors are expressed in pixels. Also, as specified in the algorithm outline (Section \ref{sec:algo}), tilde values represent estimations prior to the bundle adjustment procedure, and hat values denote estimations refined by the bundle adjustment. The first row corresponds to Step (iii) of the algorithm, the pose estimation between $C_{1}^l$ and $C_{2}^l$. The output values are consistent with the actual location of the cameras; the large angle displacements emphasize the difficulty of the task, and explain as well the limitation of the automated matching procedure in this case. 

We have also refined the relative positions of the cameras within the individual rigs. These values are provided by the stereo calibration procedure, and we validated them by performing SIFT matching between the large and small FOV cameras, and by using the stereo calibration pose as an initializer for the bundle adjustment (rows 2 and 3 in Table \ref{numbers}). The threshold for uniqueness filtering has been set as $\tau = 0.3$. However, the stereo calibration performed on site could not be done in optimal conditions. As an alternative solution, we used as pose initializations values that we obtained in the same way as for the first row of Table \ref{numbers}, by estimating and decomposing the fundamental matrix. The solutions obtained are presented on rows 4 and 5 in Table \ref{numbers}. These solutions were more accurate, and finally they have the advantage of requiring only the intrinsic camera parameters. It is worth noting that the stereo baseline is approximately 6 cm, while the distance to the scene is three orders of magnitude higher, and in these circumstances the relative angles and not the camera center relative positions will be the most relevant for scene based estimation.

\begin{table*}[htbp]
\caption{ Relative poses between analysis cameras placed on different rigs (first row), and between cameras placed on the same rig (rows 2-5). The Euler angles are expressed in degrees, and the mean reprojection errors in pixels. Tilde values represent estimations prior to the bundle adjustment procedure, and hat values denote estimations refined by the bundle adjustment. The difference between the rows 2-3 and 4-5 consists in the initialization of the bundle adjustment; in the first case we use the stereo calibration, whilst in the second case we use directly the images, in the same way as for the first row initialization. 
}
\begin{adjustwidth}{-1in}{-1in}% adjust the L and R margins by 1 inch
\begin{center}
{\footnotesize
\begin{tabular}{c|c|c|c|c|c|c|c}
{\bfseries Cam. pair } & $( \tilde{\bm{\psi}}; \tilde{\bm{\theta}}; \tilde{\bm{\phi}}) $  & $\tilde{\mathbf{C}}$ & $ \tilde{\bm{\epsilon}}$ & $( \hat{\bm{\psi}}; \hat{\bm{\theta}}; \hat{\bm{\phi}}) $  & $\hat{\mathbf{C}}$ & $ \hat{\bm{\epsilon}}$ & {\bfseries Observations }\\ 

\hline 
\hline
$C_{1}^l \Rightarrow C_{2}^l$ & $ \left (
      \begin{array}{c}
            53.27 \\
            71.52 \\
            32.94
      \end{array}
      \right ) $ & 
      $\left (
      \begin{array}{c}
            -0.78 \\
            -0.19 \\
             0.59
      \end{array}
      \right ) $ & 4.00 & $ \left (
      \begin{array}{c}
            59.68 \\
            69.11 \\
            42.75
      \end{array}
      \right ) $ & 
      $\left (
      \begin{array}{c}
            -0.81 \\
            -0.18 \\
             0.55
      \end{array}
      \right ) $ & 0.25 & Manual Init.  \\
\hline 
$C_{1}^l \Rightarrow C_{1}^s$ & $ \left (
      \begin{array}{c}
            -0.37 \\
            -0.58 \\
            0.51
      \end{array}
      \right ) $ & 
      $\left (
      \begin{array}{c}
            0.79 \\
            0.00\\
            -0.62
      \end{array}
      \right ) $ & 1.017 & $ \left (
      \begin{array}{c}
            -0.33 \\
            -0.34 \\
            0.48
      \end{array}
      \right ) $ & 
      $\left (
      \begin{array}{c}
            0.11 \\
            -0.01 \\
            -0.99
      \end{array}
      \right ) $ & 0.076 & Stereo Calib. Init.   \\
\hline 
$C_{2}^l \Rightarrow C_{2}^s $ & $ \left (
      \begin{array}{c}
            -0.19 \\
            0.72 \\
            0.23
      \end{array}
      \right ) $ & 
      $\left (
      \begin{array}{c}
            0.94 \\
            0.05\\
            -0.34
      \end{array}
      \right ) $ & 1.661 & $ \left (
      \begin{array}{c}
            -0.09 \\
            0.71 \\
            0.12
      \end{array}
      \right ) $ & 
      $\left (
      \begin{array}{c}
            0.57 \\
            0.23 \\
            0.78
      \end{array}
      \right ) $ & 0.252 & Stereo Calib. Init.   \\
\hline 
$C_{1}^l \Rightarrow C_{1}^s$ & $ \left (
      \begin{array}{c}
            -0.36 \\
            -0.23 \\
            0.43
      \end{array}
      \right ) $ & 
      $\left (
      \begin{array}{c}
            0.01 \\
            -0.06\\
             0.99
      \end{array}
      \right ) $ & 0.096 & $ \left (
      \begin{array}{c}
            -0.33 \\
            -0.28 \\
            0.47
      \end{array}
      \right ) $ & 
      $\left (
      \begin{array}{c}
            0.06 \\
            -0.02 \\
            -0.99
      \end{array}
      \right ) $ & 0.084 & SIFT Matching   \\
\hline 
$C_{2}^l \Rightarrow C_{2}^s $ & $ \left (
      \begin{array}{c}
            -0.16 \\
            0.74 \\
            0.10
      \end{array}
      \right ) $ & 
      $\left (
      \begin{array}{c}
            -0.02 \\
            -0.01\\
            -0.99
      \end{array}
      \right ) $ & 0.097 & $ \left (
      \begin{array}{c}
            -0.13 \\
            0.75 \\
            0.10
      \end{array}
      \right ) $ & 
      $\left (
      \begin{array}{c}
            -0.01 \\
            0.00 \\
            -0.99
      \end{array}
      \right ) $ & 0.087 & SIFT Matching   \\

\hline
\end{tabular}
 } %for small
\end{center}
\end{adjustwidth}
\label{numbers}
\end{table*}

\begin{figure*}[thbp]
 \begin{center}

 \begin{minipage}{0.24\textwidth}\centering
  \includegraphics[width=\textwidth]{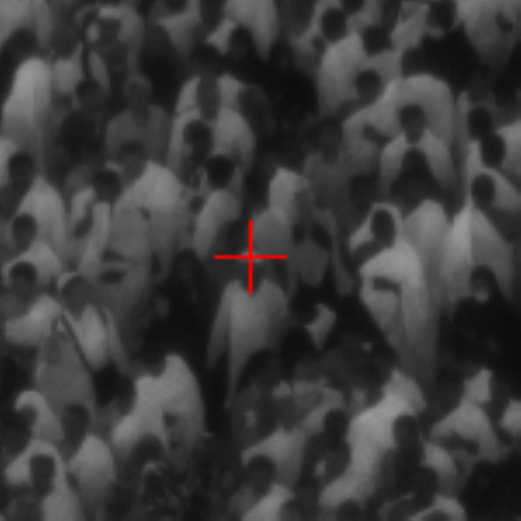}\\ 
  \includegraphics[width=\textwidth]{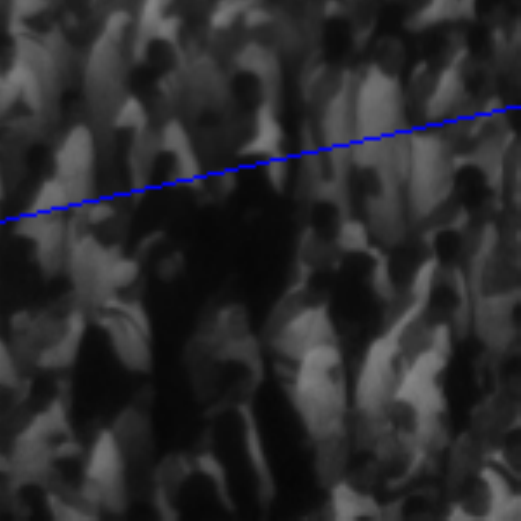}\\
  {\small a) }
 \end{minipage} 
 \begin{minipage}{0.24\textwidth}\centering
  \includegraphics[width=\textwidth]{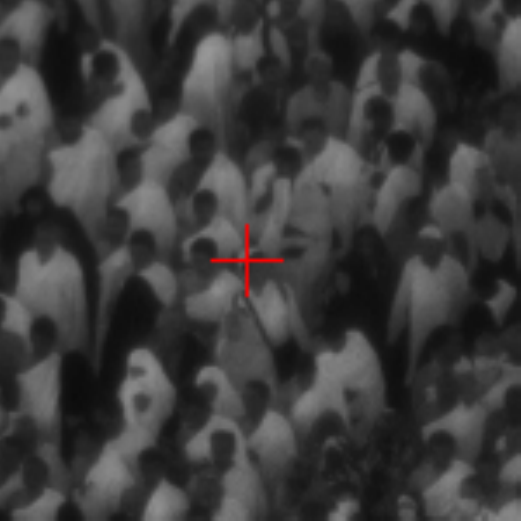}\\ 
  \includegraphics[width=\textwidth]{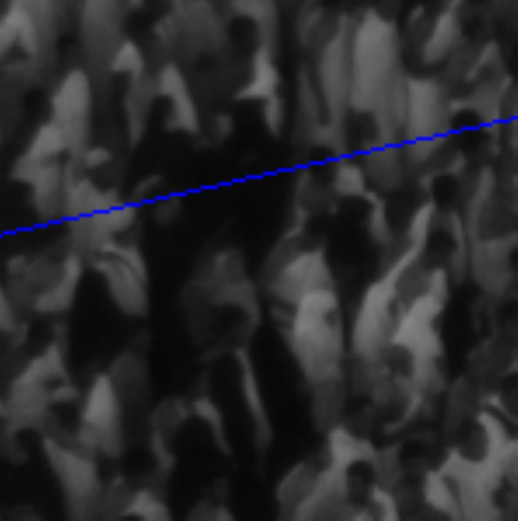}\\
  {\small b) }
 \end{minipage} 
 \begin{minipage}{0.24\textwidth}\centering
  \includegraphics[width=\textwidth]{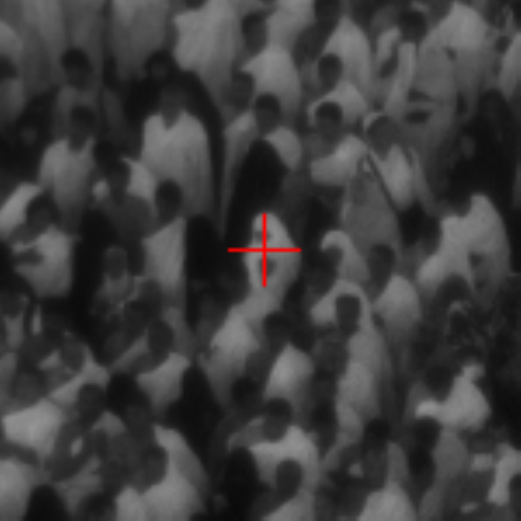}\\ 
  \includegraphics[width=\textwidth]{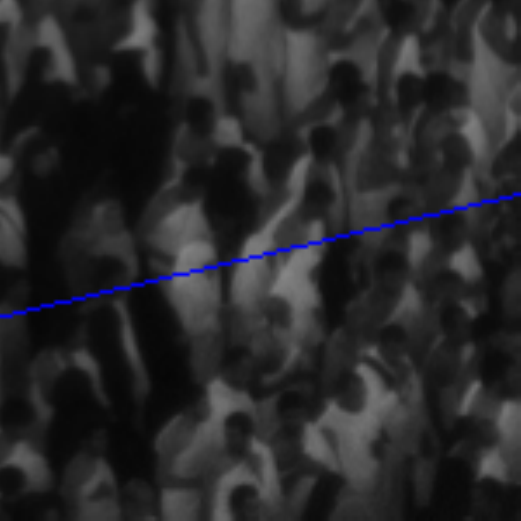}\\
  {\small c) }
 \end{minipage} 
 \begin{minipage}{0.24\textwidth}\centering
  \includegraphics[width=\textwidth]{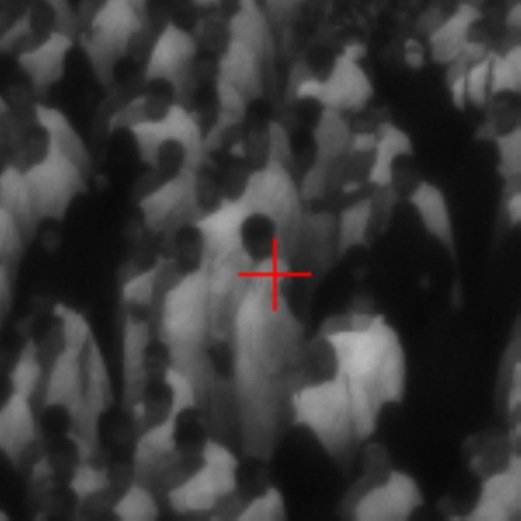}\\ 
  \includegraphics[width=\textwidth]{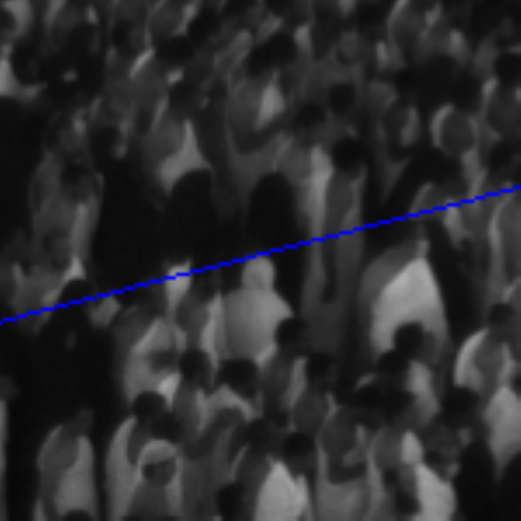}\\
  {\small d) }
 \end{minipage} 
 \\
 \begin{minipage}{0.24\textwidth}\centering
  \includegraphics[width=\textwidth]{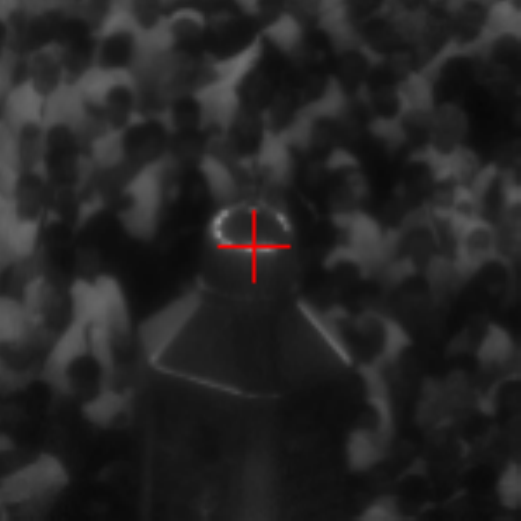}\\ 
  \includegraphics[width=\textwidth]{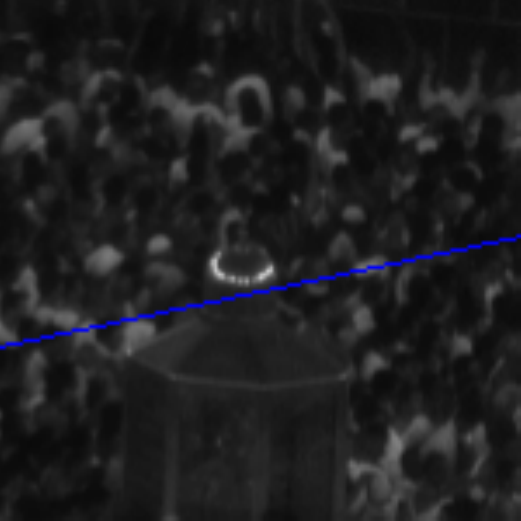}\\
  {\small e) }
 \end{minipage} 
 \begin{minipage}{0.24\textwidth}\centering
  \includegraphics[width=\textwidth]{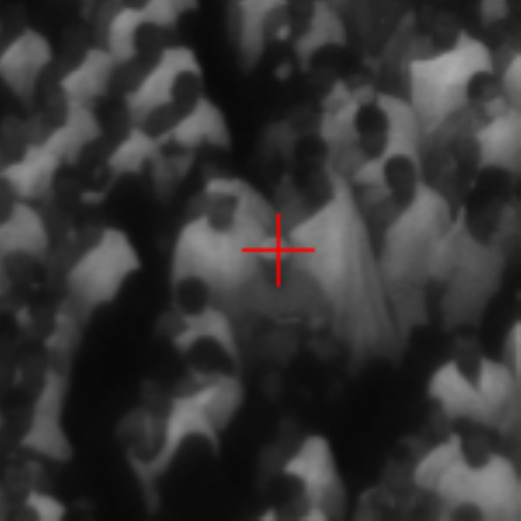}\\ 
  \includegraphics[width=\textwidth]{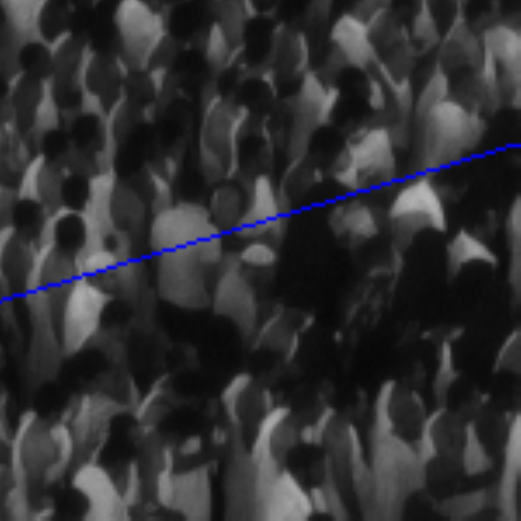}\\
  {\small f) }
 \end{minipage} 
 \begin{minipage}{0.24\textwidth}\centering
  \includegraphics[width=\textwidth]{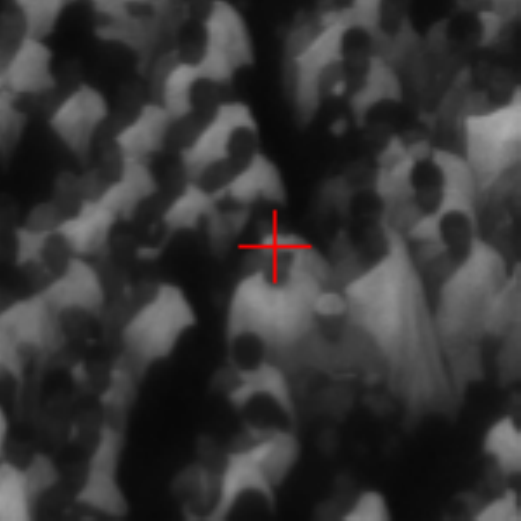}\\ 
  \includegraphics[width=\textwidth]{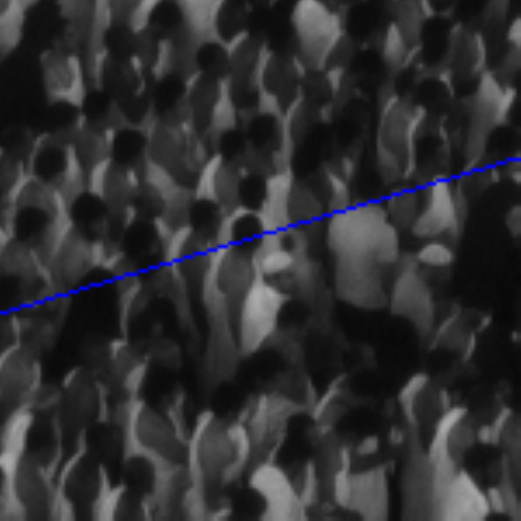}\\
  {\small g) }
 \end{minipage} 
 \begin{minipage}{0.24\textwidth}\centering
  \includegraphics[width=\textwidth]{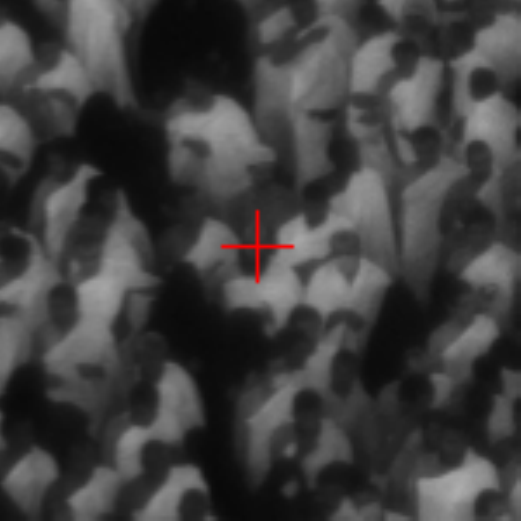}\\ 
  \includegraphics[width=\textwidth]{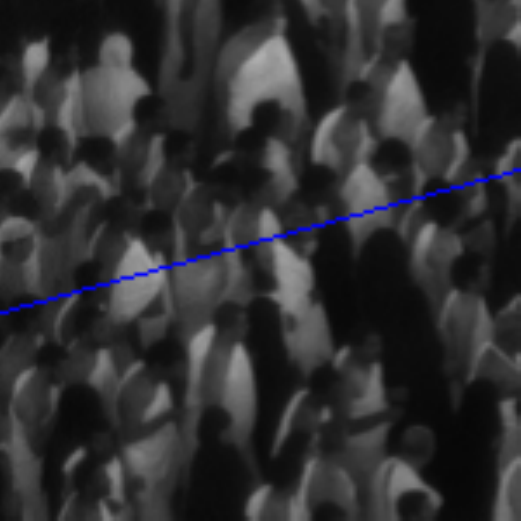}\\
  {\small h) }
 \end{minipage} 
 \end{center}
 \caption{\small A number of pixel-epipolar line correspondences between the two analysis cameras presented in Figure \ref{fig:multiM}b) and \ref{fig:multiM}d). Ideally, the correspondent of a point highlighted by the red cross in the upper row should be situated along the blue epipolar line visible in the lower row image. These results are discussed in Section \ref{subsec:disc}.}
 \label{fig:smallRes}
\end{figure*}

Having thus obtained all the necessary relative poses (rows 1, 4 and 5 in Table \ref{numbers}), we are able to estimate the relative pose between the long focal cameras. Ground truth estimations are not possible, but in order to estimate the accuracy of the result we have located in the crowd a number of salient elements (either distinctive heads, or distinctive configurations of people) and we illustrate the result by drawing for each feature the epipolar line, and judging by its proximity to the corresponding feature in the other image. In Figure \ref{fig:smallRes}, the upper row corresponds to elements identified in $C_{1}^s$ (Figure \ref{fig:multiM}b) and the lower row presents the same elements identified in $C_{2}^s$ (Figure \ref{fig:multiM}d). The following remarks are necessary at this point. Firstly, the perspective change makes the correspondence search very tedious even for a human. Secondly, the drawing of the epipolar line has actually assisted us in pinpointing most of these correspondences, and we are confident that the method will be helpful in automating these tasks.

\subsection{Discussion of the dense crowd results}
\label{subsec:disc}
Overall, the distance in the image space between the corresponding element and the corresponding epipolar line is in the range of a few pixels. The major factors responsible for these misalignments are the inaccuracies in estimating the intrinsic parameters, as well as the errors related to the relative pose estimations - but for the dimensions of the scene involved in the experiment, we argue that the results are very promising. 

Moreover, some areas of the scene exhibit near perfect alignments. The first four matches presented in Figure \ref{fig:smallRes}:
\begin{itemize}
\item the white cap man in a) positioned under the epiline, in the left part of the patch;
\item same for b), the person looking slightly towards the left;
\item the woman in c) wearing a white veil, and positioned in front of two other women similarly dressed;
\item the woman in d) wearing a white veil, and positioned with the back towards the second camera;
\end{itemize}
are very accurate, in spite of the fact that in one of the images the first three persons are located near the border, a fact which potentially increases radial distortion related errors. 

We could also identify the following correspondences which exhibit small but visible misalignments:
\begin{itemize}
\item the shiny circumference of the Station of Ibrahim, depicted in e)
\item another two men wearing white caps, presented in f) and g)
\item a distinctively bearded man presented in h)
\end{itemize}

The fact that the epipolar line does not pass precisely through the corresponding element is not detrimental for the purpose of association and tracking in the crowd. Assuming that the person is not occluded, using this extra information we would not only be able to trim down the research space to a band along the epipolar line, but also if we were able to position the ground plane within the same coordinate system we would further reduce the research space to a fraction of the band. Of course, in order to do dense matching reliably we still need a neighborhood based similarity measure that has to be resilient to major perspective change and occlusions; this is a promising direction of research that we intend to follow in order to benefit from the relative pose algorithm we propose, and ultimately in order to perform dense associations. 

Finally, the present results of the proposed algorithm on this type of data are also encouraging as they illustrate the potential of multi-camera systems in extremely crowded environments. In the current research context, this application field has been associated mostly with single camera systems \citep{Wang13}, but paradoxically it would greatly benefit from multi-view systems given the frequent occlusions and scene clutter that characterize it.

\section{Conclusion}
\label{sec:conclusion}

In this paper we propose a new method for aligning multiple cameras analysing a homogeneous scene. Our method addresses the settings where for practical reasons calibration pattern/object based registrations are not possible. By employing stereo rigs featuring a long focal analysis camera and a short focal registration camera, the proposed solution alleviates the requirement to get access to the studied scene. The fact that we are using a large FOV simultaneously allows us to avoid making any assumptions about the homogeneous region we analyse, such as the presence of shades, silhouettes etc. A first experiment has been conducted in an indoor environment and has shown, by using interest point correspondences in the analysis area as ground truth, that this method can guide the relative pose estimation for scenes poor in salient features in a coarse-to-fine manner supported by hardware. The second test has shown that in the absence of any salient features, the method is capable of providing a full calibration of the analysis cameras in a difficult, large scale scenario. 

In the future, we would like to investigate the applicability of the proposed hybrid stereo solution in two frequently recurring settings. We intend to employ this method as a preprocessing step for a wide range of homogeneous pattern analysis applications, such as those related to the extraction of accurate models for highly dense crowd dynamics. Secondly, we would like to evaluate further the potential of this solution in specific applications such as autonomous robot navigation or image alignment and stitching, which employ pyramid based coarse-to-fine optimizations; our setup augments these systems by supplementing the image pyramid with a level provided by an independent data source.

\begin{acknowledgements}
This work was funded by the Qatar National Research Fund (part of the Qatar Foundation) under the grant NPRP 09-768-1-114. The authors acknowledge A. Gutub and his team at the Centre of Research Excellence in Hajj and Omrah; and O. Gazzaz, F. Othman, A. Fouda and B. Zafar at the Hajj Research Institute (Umm al-Qura university, Makkah KSA) for their organisation and logistical support in the video data collection at the grand mosque (Al-Harram) in Makkah, as well as for useful discussions. The authors would also like to thank F. Ali for his help in data collection, and R. Saussard for his assistance in the coding of some of the algorithms. Lastly, the authors would like to thank the late Professor Maria Petrou who was a co-investigator on this project and without whose consistent support the project would not have been possible -- this article is dedicated to her memory. 
\end{acknowledgements}

% BibTeX users please use one of
%\bibliographystyle{spbasic}      % basic style, author-year citations
\bibliographystyle{spmpsci}      % mathematics and physical sciences
%\bibliographystyle{spphys}       % APS-like style for physics
%\bibliography{bibfovstereo}   % name your BibTeX data base

% Non-BibTeX users please use
% \begin{thebibliography}{}
% %
% % and use \bibitem to create references. Consult the Instructions
% % for authors for reference list style.
% %
% \bibitem{RefJ}
% % Format for Journal Reference
% Author, Article title, Journal, Volume, page numbers (year)
% % Format for books
% \bibitem{RefB}
% Author, Book title, page numbers. Publisher, place (year)
% % etc
%\end{thebibliography}

\end{document}